\documentclass[cameraready]{Interspeech}
\usepackage{CJKutf8}
\usepackage{siunitx}
\usepackage{subcaption}
\usepackage{cite}
\usepackage{hyperref}
\usepackage{cleveref}
\title{Prosodic ABX: A Language-Agnostic Method for Measuring\\ Prosodic Contrast in Speech Representations}

\author[affiliation={1}, equalcontribution, orcid=0009-0008-5134-1936]{Haitong}{Sun}
\author[affiliation={1}, equalcontribution, correspondingauthor, orcid=0009-0009-4868-2868]{Stephen}{McIntosh}
\author[affiliation={2}, orcid=0000-0001-5254-1093]{Kwanghee}{Choi}
\author[affiliation={2}, orcid=0009-0003-4452-4084]{Eunjung}{Yeo}
\author[affiliation={1}, orcid=0000-0002-6265-9674]{\\Daisuke}{Saito}
\author[affiliation={1}, orcid=0000-0002-8778-9555]{Nobuaki}{Minematsu}
\address{
    $^1$ The University of Tokyo, Japan \quad $^2$ University of Texas at Austin, USA
}

\email{\{sunhaitong,smcintosh,mine\}@gavo.t.u-tokyo.ac.jp}

\keywords{prosody, self-supervised learning, model analysis, comparative assessment, stress, pitch accent, tone}

\usepackage{comment}

\begin{document}

\maketitle

\begin{abstract}
    Speech representations from self-supervised speech models (S3Ms) are known to be sensitive to phonemic contrasts, but their sensitivity to prosodic contrasts has not been directly measured.
    The ABX discrimination task has been used to measure phonemic contrast in S3M representations via minimal pairs. We introduce prosodic ABX, an extension of this framework to evaluate prosodic contrast with only a handful of examples and no explicit labels. Also, we build and release a dataset of English and Japanese minimal pairs and use it along with a Mandarin dataset to evaluate contrast in English stress, Japanese pitch accent, and Mandarin tone. Finally, we show that model and layer rankings are often preserved across several experimental conditions, making it practical for low-resource settings.
\end{abstract}

\section{Introduction}
Speech representations from hidden layers of S3Ms such as wav2vec 2.0~\cite{baevski2020wav2vec} and HuBERT~\cite{hsu2021hubert} are widely used due to their sensitivity to linguistic information~\cite{wells2022phonetic, mohamed2022selfsupervised}. Discretizations of these representations (\textit{S3M tokens}) are known to be phonetic~\cite{baevski2020wav2vec, hsu2021hubert} and have enabled the creation of language models trained entirely on speech~\cite{lakhotia2021generative}.

Prosody refers to the elements of human speech that extend across phonetic boundaries, such as stress, intonation, and rhythm. We use it for speech segmentation~\cite{mehler1981syllables}, word distinction, syntactic ambiguity resolution~\cite{kjelgaard1999prosodic}, focus~\cite{bolinger1972accent}, and more~\cite{cutler1997prosody, dahan2015prosody}.
This raises a question: how much prosodic information is contained in S3M representations?

It has been shown that S3M representations are sensitive to artificial manipulation of pitch and intensity~\cite{onda2025benchmarking} and can be used to predict acoustic correlates such as pitch~\cite{lin2023utility}. Prosodic categories are also encoded: Bentum et al.~\cite{bentum2025word} show that stress can be best recovered from XLSR~\cite{babu2022xlsr} representations at late middle layers and Shen et al.~\cite{shen2024encoding} show that pretraining language may affect encoding of lexical tone and that it improves after ASR fine-tuning. Finally, de la Fuente et al.~\cite{de2024layer} find that while layer-wise curves have different shapes, peak classification accuracy on stress and tone is similar across wav2vec 2.0 and HuBERT-style models.

Probing studies show that prosodic information \textbf{exists} in S3M representations, but not that it is \textbf{prominent} in representation space~\cite{lin2023utility,choi2024understanding}. Prominence is particularly important for distance-based applications such as clustering~\cite{hsu2021hubert} and utterance comparison~\cite{bartelds2022neural}, where meaningful contrasts should be directly reflected in representation geometry.

To examine this property in S3Ms, we adapt the ABX task to evaluate prosodic contrast. Originally conceived to evaluate whether humans can perceive the differences between two stimuli, the ABX task was adapted by Schatz et al.~\cite{schatz2013evaluating} to measure phonemic contrast in machine representations via minimal pairs like \textit{beg} and \textit{bag}. In this framework, two recordings of the same category should be closer in representation space than recordings of different categories. 

\begin{figure}[t]
    \centering
    \begin{subfigure}{\linewidth}
        \centering
        \includegraphics[width=0.9\linewidth]{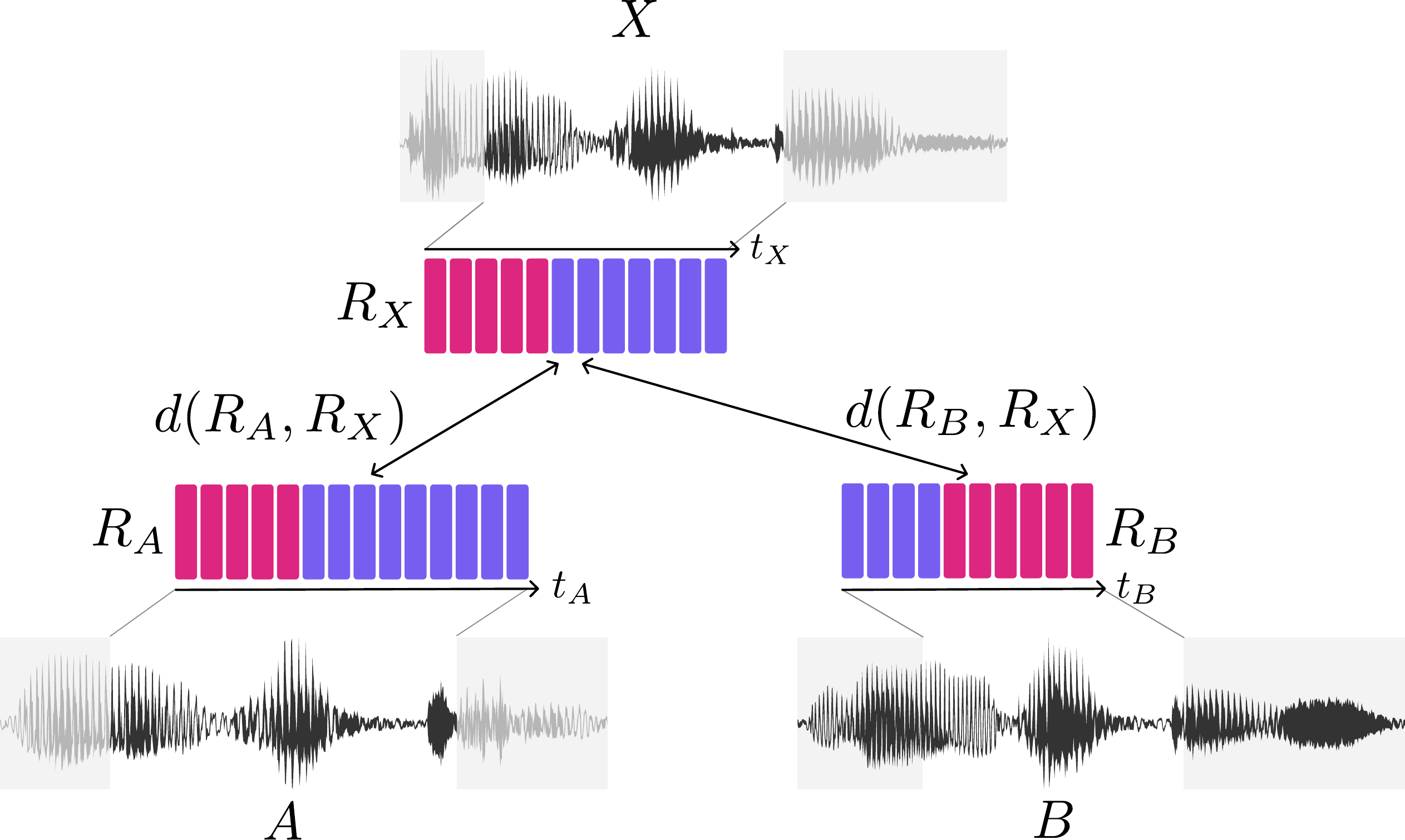}
        \caption{\textbf{Conceptual overview.} $A$, $B$, and $X$ have the same phonemic sequence, but $B$ has a different prosodic pattern. This is illustrated by the colors of the representations, $R_A$, $R_B$, and $R_X$. We align $R_A$ and $R_B$ with $R_X$ using dynamic warping and check whether the alignment cost (or distance) $d$ to $R_X$ is smaller for $R_A$ than for $R_B$.}
        \label{fig:method}
    \end{subfigure}

    \begin{subfigure}{\linewidth}
    \vspace{1em}
    \centering
        \includegraphics[width=0.9\linewidth]{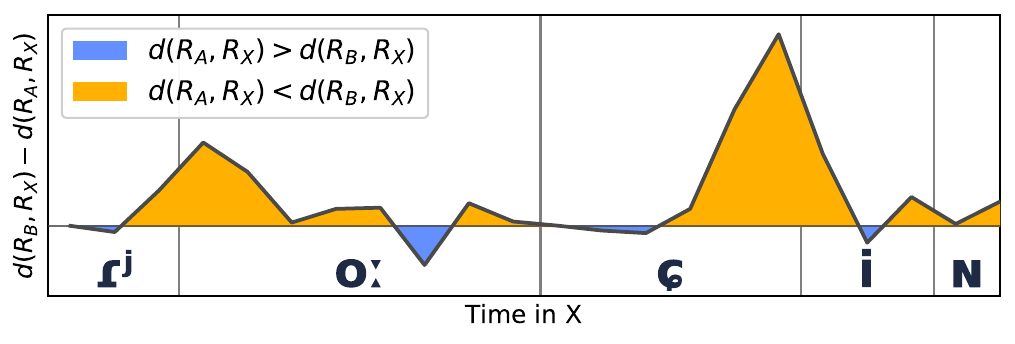}
        \caption{\textbf{Difference in local DTW distance $d$ for a Japanese pitch-accent ABX triplet.} Here, $A$ and $X$ correspond to \textipa{\textbf{/R\super j\'o:}.Cin/} (initial-accented), while $B$ corresponds to \textipa{/R\super jo:.Cin/} (unaccented). The distance used in prosodic ABX is the normalized area under this curve.}
        \label{fig:ryoshin_dtw_japanese-hubert}
    \end{subfigure}
    \vspace{-1.5em}
    \caption{\textbf{Our prosodic ABX framework.}}
    \vspace{-1em}
\end{figure}

ABX also has several practical advantages compared to categorical probing. First, probing requires labeled datasets, which can be difficult and time-consuming to construct, especially for languages with complex prosodic patterns. It also requires labeled data to train a classifier. In contrast, the ABX test evaluates representations directly and uses the entire dataset without additional training, simplifying data preparation. Second, categorical labels may not directly reflect the contrasts that listeners rely on for comprehension. By using minimal pairs, ABX enables more naturalistic evaluation and facilitates direct comparison with human perceptual judgments~\cite{millet2020perceptimatic, dunbar2022selfsupervised}.

In addition, ABX compares representations using dynamic time warping (DTW), whereas probing methods typically aggregate variable-length representations via mean pooling. Mean pooling collapses the temporal structure of a sequence, which can obscure time-varying prosodic patterns. In contrast, DTW aligns two sequences while allowing local temporal variation, making it well-suited for comparing prosodic contours across utterances (see Figure~\ref{fig:ryoshin_dtw_japanese-hubert}). These properties are important for interpretable speech comparison, where prosodic ABX can help determine whether a given representation is sensitive to relevant prosodic contrasts, and its time alignments show where in the utterance each contrast is encoded. Such analyses can also support applications such as visual pronunciation feedback to language learners~\cite{mcghee2025comparative, minematsu2023unified}.

In this paper, we extend the ABX framework to prosodic minimal pairs (see Figure~\ref{fig:method}), enabling training-free analysis of lexical prosody encoding in S3Ms. We evaluate this approach on three languages with distinct lexical prosodic systems: English (lexical stress), Japanese (pitch accent), and Mandarin (lexical tone).
Our contributions are threefold.
\textbf{First}, we propose a prosodic ABX framework to measure prosodic contrast in S3M representations.
\textbf{Second}, we collect a dataset of labeled stress and pitch accent minimal pairs in English and Japanese, respectively. To our knowledge, this is the first publicly available dataset that provides clean recordings of prosodic minimal pairs in English and Japanese extracted from read sentences.
\textbf{Finally}, we apply our framework to compare human listeners and a wide selection of S3Ms. We further show that results on S3Ms are consistent under various experimental conditions, demonstrating that the framework can be applied even for languages where data is limited or unavailable.

\section{Prosodic ABX} \label{sec:method}
Prosodic ABX is a training-free method that evaluates the extent to which speech representations emphasize linguistically relevant prosodic features. Like the standard ABX task, we prepare a \textit{triplet} of speech samples $A$, $B$, and $X$. $A$ and $B$ share the same phonemic sequence and speaker, but have a different prosodic pattern that generates semantic contrast. These constitute a prosodic \textit{minimal pair}.\footnote{By using semantically contrastive minimal pairs, both members of the pair can be naturally elicited from the same speaker in appropriate semantic contexts, simplifying the data collection process.}  For example, we might have $A =\ $ \textbf{sub}ject (noun, \textipa{/"s\textturnv b.\t{dZ}ekt/}) and $B =\ $ sub\textbf{ject} (verb, \textipa{/s\textturnv b"\t{dZ}ekt/}). $X$ is another sample from a \textit{different} speaker with the \textit{same} phonemic content and prosody pattern as $A$.

We feed these samples to the model and obtain representation sequences $R_A$, $R_B$, and $R_X$ of lengths $t_A$, $t_B$, and $t_X$, respectively.
To reduce the effects of correlated context, we clip minimal pairs from the surrounding audio before encoding~\cite{choi2024understanding,choi2024selfsupervised}. In \Cref{sec:in-context-results}, we analyze results without this clipping step.

We then compare representation sequences using dynamic time warping (DTW). DTW aligns two sequences while allowing local temporal variation, minimizing the total frame-wise distance along the alignment path. This produces both an overall DTW distance $d_\text{raw}$ and a path that localizes the contributions to that distance over the length of the input (as in \Cref{fig:ryoshin_dtw_japanese-hubert}). To account for differences in sequence length, we divide $d_\text{raw}$ by the path length to obtain a normalized distance $d$.

If $d(R_A, R_X) < d(R_B, R_X)$, then this is evidence that the model correctly distinguishes the minimal pair. So the \textit{ABX score} \cite{poli2025fastabx} for a triplet is calculated as:
\begin{align}
S(A, B, X) = \begin{cases} 1 & \text{if } d(R_A, R_X) < d(R_B, R_X) \\ 0.5 & \text{if } d(R_A, R_X) = d(R_B, R_X) \\0 & \text{otherwise} \end{cases}
\end{align}

To estimate the probability of detecting a particular type of prosodic contrast, we calculate the ABX score on a dataset of minimal pairs. We score each triplet, then perform nested averaging: first over speaker pairs, producing one score per contrast, and then over all contrasts.  We subtract this ABX score from $1$ to report it as an \textit{error rate}. This procedure follows phonemic ABX~\cite{schatz2013evaluating}, except that instead of comparing phones within a fixed phonemic context, we compare prosodic contours realized over identical phonemic content.\footnote{Our implementation is available at \url{https://github.com/stephenmac7/prosodic-abx}, which uses the \texttt{fastabx} library~\cite{poli2025fastabx} originally developed for phonemic ABX.}

\section{Prosodic Minimal Pair Datasets}
We constructed prosodic minimal pair datasets for three languages representing distinct lexical prosodic systems: English, Japanese, and Mandarin. For English and Japanese, we collected prompted recordings from native speakers, while for Mandarin we used recordings from an existing corpus. \Cref{tab:dataset_stats} summarizes the minimal pair datasets used in our experiments.

\subsection{Prosodic Minimal Pair Selection}\label{ssec:pair-selection}
First, we selected minimal pairs that differ only in their lexical prosodic pattern while sharing the same phonemic sequence.

\noindent \textbf{English lexical stress pairs.} English words have one syllable with primary stress, usually marked by higher pitch, longer duration, greater intensity, or a shift in vowel quality~\cite{vanheuven2018acoustic}. We selected 15 noun-verb pairs 
from the CMU Pronouncing Dictionary with identical phoneme sequences but different primary stress positions, where 4 exhibited vowel weakening.

\noindent \textbf{Japanese pitch accent pairs.} Japanese words are associated with characteristic pitch patterns that may distinguish lexical meaning (e.g., \textipa{/\textbf{\'a}.me/} ``rain''--\textipa{/a.me/} ``candy''). We selected 23 minimal pairs with the same kana\footnote{Kana is a phonemic script in which each character represents a \textit{mora}, an equal-length timing unit. Morae are no longer than syllables.} sequence but different pitch accents from lists provided by Japanese language educators. Only words between 2 and 4 morae long were selected.

\noindent \textbf{Mandarin tone minimal pairs.} In Mandarin, syllables are distinguished by lexical tones, realized as distinct pitch contours (e.g., \textipa{/ma\textbf{\tone{55}}/} ``mother''--\textipa{/ma\textbf{\tone{214}}/} ``horse''). We selected 2310 pairs from 385 monosyllable pinyin sequences. These samples are drawn from the MCAE-Monosyllable corpus~\cite{li2025mandarin}, which contains recordings of 422 monosyllables produced by six professional Mandarin actors with the four lexical tones. As not all pinyin–tone combinations are available for every speaker, only syllables where each tone is produced by at least 3 speakers were included.

\subsection{Prompted reading corpus} \label{ssec:prompted-corpus}

\textbf{English corpus.} 
    For each noun-verb minimal pair, we constructed four carrier sentences, two targeting the noun usage (e.g., \textit{A \textbf{per}mit is required for long stays.}) and two targeting the verb usage (e.g., \textit{Some hotels per\textbf{mit} smoking.}). This resulted in a total of 60 sentences, corresponding to two recording sets for each minimal pair.
    The corpus was recorded in a studio by 10 native speakers of American English. Since there is variation among American English speakers on even these noun-verb pairs, the stress of each target word was manually verified by a native speaker.

\noindent \textbf{Japanese corpus.} 
    For each pitch-accent minimal pair, we constructed two pairs of carrier sentences with identical context for both words. The corpus was recorded in a studio by 10 native speakers, then each sample was reviewed by a native speaker.

English and Japanese corpora were aligned with the Montreal Forced Aligner (MFA)~\cite{mcauliffe2017montreal} to obtain the boundaries of the target word. These timestamps were reviewed and adjusted by native speakers familiar with phonology.\footnote{The dataset is publicly available at \url{https://huggingface.co/datasets/HaitongSUN/prosodic-abx}.}

\subsection{Synthesized corpus}

We additionally synthesized all prosodic minimal pairs using TTS to evaluate whether synthesized speech can provide a cost-effective proxy for natural speech for prosodic ABX (\Cref{synth-results}).
We synthesized speech for all three languages using the Google Cloud Text-to-Speech (G-TTS) API~\cite{google_tts}. We used the standard voices, which support control of prosodic features and produce consistent speech. For each language, we selected four voices.

We also synthesized the English stress minimal pairs using Kokoro~\cite{kokoro_tts}, a neural open-weight TTS model that provides explicit control over phonetics and stress. Since Kokoro is unstable for very short utterances, we synthesized examples using a constant carrier sentence and extracted timestamps using MFA~\cite{mcauliffe2017montreal}. 
We include this additional synthesis condition to examine how conclusions drawn from evaluation on synthetic speech may depend on the type of synthesizer.

% \vspace{0.5\baselineskip}

\noindent

% generated with scripts/format_dataset_stats_table.py (manually edited 46 -> 23 because the Japanese ABX only compared identical context so the reported phone sequence is actually kana_context#
\begin{table}[t]
  \caption{\textbf{Overview of the prosodic minimal pair datasets.} Duration (in minutes) includes the minimal pairs only.}
  \label{tab:dataset_stats}
  \centering
  \small
  \resizebox{\columnwidth}{!}{%
  \begin{tabular}{l c c c c}
    \toprule
    \textbf{Language} & \textbf{Source} & \textbf{\#Pairs} & \textbf{\#Spks/Voices} & \textbf{Duration} \\
    \midrule
    English & Recording & 15 & 10 & 4.6 \\
     & G-TTS & 15 & 4 & 1.8 \\
     & Kokoro & 15 & 4 & 1.0 \\
    \midrule
    Japanese & Recording & 23 & 10 & 5.3 \\
     & G-TTS & 23 & 4 & 4.2 \\
    \midrule
    Mandarin & MCAE~\cite{li2025mandarin} & 2310 & 6 & 100.2 \\
     & G-TTS & 2310 & 4 & 65.5 \\
    \bottomrule
  \end{tabular}
  }
\end{table}

\section{Experiments}

\subsection{Speech representations}

We evaluated a diverse set of S3Ms.
We considered models based on wav2vec~2.0~\cite{baevski2020wav2vec} and HuBERT~\cite{hsu2021hubert}, which represent two major paradigms: contrastive learning and masked prediction, respectively. For both architectures, we included base and large model variants, as well as models pretrained on English \cite{baevski2020wav2vec,hsu2021hubert}, Japanese \cite{reazon-research-japanese-wav2vec2-base,reazon-research-japanese-wav2vec2-large,reazon-research-japanese-hubert-base-k2,k2ssl,rinna-japanese-hubert-large,sawada2024release}, and Mandarin \cite{tencentgamemate}. We also included XLSR-53~\cite{babu2022xlsr} and mHuBERT-147~\cite{boito2024mhubert147}, which were trained on multilingual corpora, and WavLM~\cite{chen2022wavlm}, a widely-adopted denoising masked prediction model. Overall, we selected 17 S3Ms for the experiments, and for each performed the prosodic ABX for all layers.
We compared S3Ms with two acoustic baselines: mel spectrograms and MFCCs.

\subsection{Human experiment}

To provide human reference points, we conducted a human ABX experiment, where humans were presented with three audio stimuli $A$, $B$, and $X$ in sequence and prompted to select which of $A$ or $B$ is more similar to $X$. Human ABX experiments conducted on the English and Japanese natural recording corpora covered all prosodic contrasts, while Mandarin samples were arbitrarily selected due to dataset size.
We recruited native speakers of American English ($n=33$), Japanese ($n=18$), and Mandarin ($n=15$)\footnote{Median completion time (min): 14 (EN), 16 (JA), 23 (ZH)}. One-tenth of the triplets shown to each participant were catch trials consisting of trivial identity-matching, following Millet et al.~\cite{millet2020perceptimatic}. Participants with high catch-trial error rates were replaced.
Triplets are counterbalanced such that the correct answer is equally likely to be $A$ or $B$. No participant listened to the same recording twice.

\section{Results}
\subsection{Human vs. S3M performance on natural speech} \label{human-results}
\begin{figure}[t]
    \centering
    \includegraphics[width=\linewidth]{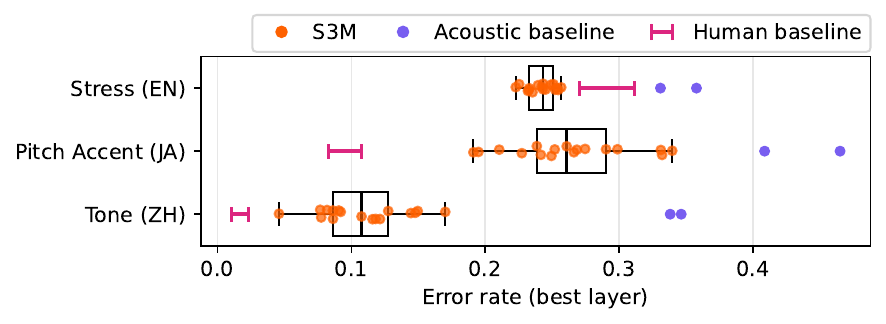} 
    \vspace{-2em}
    \caption{\textbf{ABX error rates ($\downarrow$) across different prosodic tasks.} We compare the best layer of each S3M (orange box plots, $n=17$), acoustic baselines (violet dots), and humans (pink 95\% confidence intervals).}
    \label{fig:human_ssl_boxplot}
\end{figure}

\Cref{fig:human_ssl_boxplot} compares human and S3M performance on natural speech to determine whether S3M representations are as sensitive to prosodic contrast as humans. For each S3M, we only report the highest score achieved by any layer.\footnote{The best-performing representation was Chinese HuBERT-large~\cite{tencentgamemate} layer 15, which was optimal within 0.4\% for all three languages.
For ABX error rates of individual S3Ms, see \\ \url{https://stephenmac7.github.io/prosodic-abx}}

On all tasks, S3Ms substantially outperform random chance (50\%) and the acoustic baselines, confirming that prosodic distinctions are prominent in S3M representations. While native English listeners underperform the worst S3Ms on stress (29\% vs. 26\%), native Japanese and Mandarin listeners outperform the best models on pitch accent (9\% vs 19\%) and tone (2\% vs. 5\%).

\begin{figure}[t]
  \centering
  \includegraphics[width=0.65\linewidth]{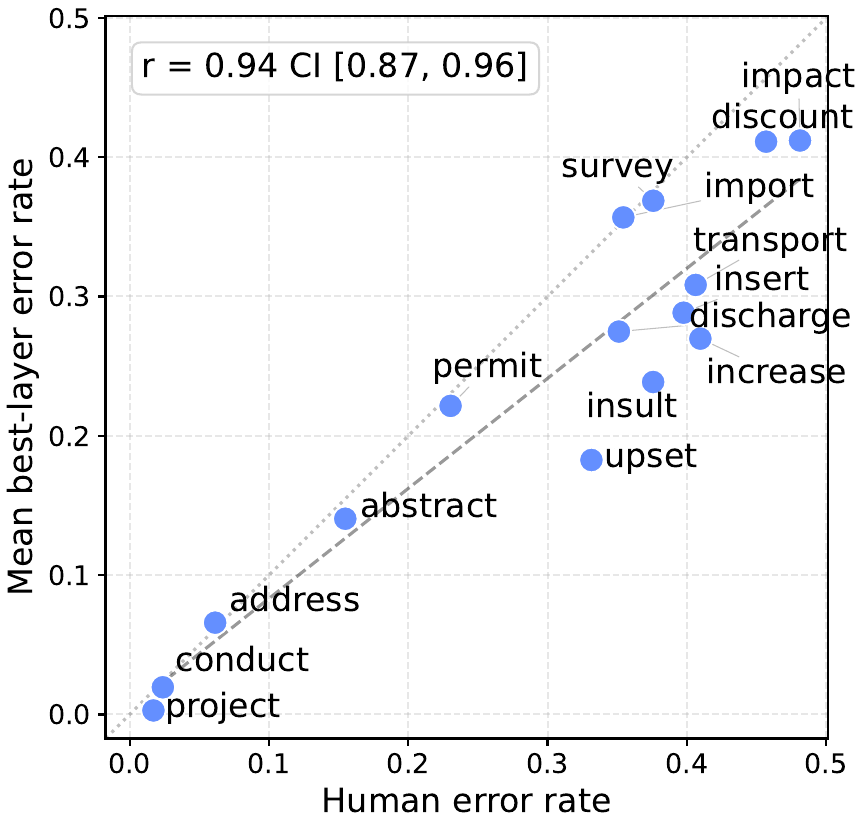}
  \vspace{-1em}
  \caption{\textbf{Word-level ABX error rates for English lexical stress: human participants vs S3Ms}.}\label{fig:human_vs_machine_word_error_English_stress}
\end{figure}

We also take a closer look at the English ABX experiments. \Cref{fig:human_vs_machine_word_error_English_stress} shows a strong word-level correlation ($r = 0.94$) between human and model error rates for English stress minimal pairs, indicating that humans and S3Ms struggle on the same kinds of pairs. Notably, both humans and models perform best on pairs with vowel weakening (``project'', ``conduct'', ``address'', and sometimes ``abstract'').

These findings suggest that for English stress, where prosodic cues are more complex and less consistently tied to semantics~\cite{vanheuven2018acoustic}, S3Ms can outperform human listeners, but they trail humans on Japanese pitch accent and Mandarin tone. Furthermore, a correlation between S3M and human error rates suggests that contrasts that are easy for humans are also easily distinguishable in S3M representations.

\subsection{Synthesized speech as a proxy} \label{synth-results}

\begin{table}[t]
\caption{\textbf{TTS proxy quality for layer and model selection.} Subscript $m$ denotes a median across models. Regret is the increase in error rate on natural speech when the best-performing layer is estimated using synthesized speech. $\rho_\textrm{model}$ is the Spearman rank correlation of best-layer error rates across models. G = G-TTS; K = Kokoro.}
\label{tab:synth-proxy}
\centering
\small
\resizebox{\columnwidth}{!}{%
\begin{tabular}{l c c @{\quad} c @{\enspace} l @{\;} r}
\toprule
 & $r_m$ & Regret$_m$ (\%) & \multicolumn{3}{c}{$\rho_\textrm{model}$ [95\% CI]} \\
\cmidrule(r){2-3} \cmidrule(r){4-6}
Japanese (G) & 0.93 & 0.04 & 0.81 & $[\phantom{-}0.53,$ & $0.93]$\\
Mandarin (G) & 0.93 & 0.07 & 0.52 & $[\phantom{-}0.10,$ & $0.89]$\\
English (G) & 0.50 & 0.76 & 0.35 & $[-0.15,$ & $0.79]$\\
English (K) & 0.85 & 0.47 & 0.61 & $[\phantom{-}0.25,$ & $0.82]$\\
\bottomrule
\end{tabular}
}
    \vspace{-1em}
\end{table}

We evaluated whether synthesized speech is a good proxy for natural speech by comparing layer-wise performance on prosodic ABX (\Cref{tab:synth-proxy}).
For each model, we measured (1) the \textit{Pearson correlation $r$} between layer-wise ABX scores on natural and synthesized speech and (2) \textit{regret}, the increase in error rate on natural speech observed 
%when choosing the best-performing layer on synthesized speech.
when the layer is selected based on performance on synthesized speech.
Layer-wise correlation is high for Japanese ($r=0.93$) and Mandarin ($r=0.93$), but lower for English ($r=0.50$ and $r=0.85$). Regret is negligible for pitch accent and tone, but non-trivial for stress, especially given the compressed distribution of English results (IQR $= 1.7\%$, see \Cref{fig:human_ssl_boxplot}).

We also examined whether synthesized speech preserves model rankings relative to natural speech. Rankings are well-preserved in Japanese ($\rho = 0.81$). While correlations are also positive in Mandarin ($\rho = 0.52$) and English (Kokoro) ($\rho = 0.61$), confidence intervals are wide, likely due to the limited number of models evaluated. For English G-TTS, the correlation is not significant.

Collectively, these results suggest that synthesized speech is a practical proxy for layer selection in tone and pitch accent tasks. Model rankings are well preserved for pitch accent, but appear less reliable in other settings.

\subsection{In-context vs. out-of-context} \label{sec:in-context-results}
For our experiments, we used minimal pairs clipped from the surrounding audio (out-of-context), since the paired samples typically occur in different contexts. However, in practice, systems operate on continuous speech rather than pre-segmented words, making in-context behavior more relevant.

To address this, we compare in-context and out-of-context ABX scores computed from S3M representations. Our Japanese dataset contains minimal pairs embedded in identical carrier sentences, allowing us to encode the full utterance and compare the frames corresponding to the minimal pair \cite{pasad2021layerwise,pasad2023comparative}.

First, we compared layer-wise prosodic ABX results between out-of-context and in-context settings. The error curves are highly correlated (per-model median $r=0.93$).
Model rankings are also highly correlated between in- and out-of-context settings ($\rho=0.91$, 95\% CI $[0.71, 0.99]$).

Second, for all tested models, in-context performance is better (median best-layer $\Delta = \text{out}-\text{in} = 9.4\%$), and this advantage grows with depth (median $\rho(\textrm{layer index}, \Delta) = 0.43$, Wilcoxon signed-rank $p=0.002$), consistent with the hypothesis that deeper layers take more advantage of context mixing.
After accounting for layer depth via linear regression, partial correlation between in- and out-of-context settings gets even higher ($r=0.97$).

In summary, models perform better on in-context speech, mirroring the way human perception is enhanced by extrinsic prosodic cues~\cite{vanheuven2018acoustic, sugiyama2021effect, huang2009general}. Moreover, out-of-context prosodic ABX is predictive of in-context layer and model rankings and can hence be used when parallel datasets are unavailable.

\subsection{ABX scores are correlated across all three contrasts} \label{cross-task}

\begin{figure}[t]
  \centering
  \includegraphics[width=1.0\linewidth]{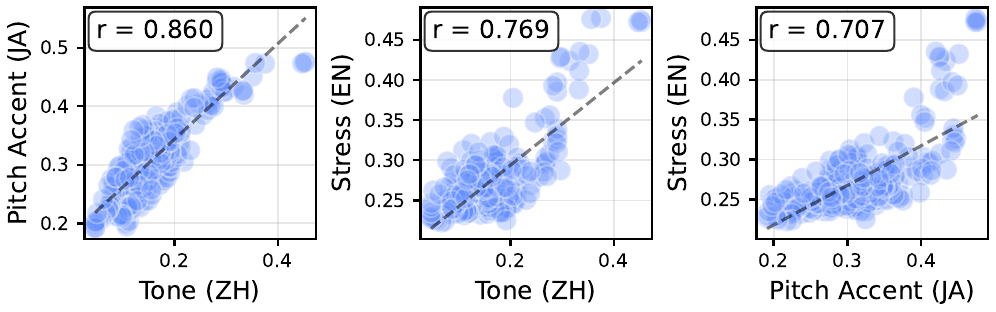}
  \vspace{-1em}
  \caption{\textbf{Error-rate correlation across prosodic tasks.} For all S3Ms, each layer is plotted according to its error rates.}
  \label{fig:cross-task-correlation}
    \vspace{-1em}
\end{figure}

To measure whether high contrast on one prosodic task may extend to others, we calculated the pairwise correlation between error rates on the three tasks (Figure~\ref{fig:cross-task-correlation}).
ABX error rates on tone and pitch accent are strongly correlated. Stress is also moderately correlated with both.
While language-specific effects have been observed in previous work~\cite{bentum2025word}, representations sensitive to one type of prosodic contrast tend to perform well on others.
We hypothesize that this is because these tasks often share acoustic correlates such as F0.

\section{Conclusion}
We proposed a prosodic ABX framework and constructed prosodic minimal pair datasets for English, Japanese, and Mandarin.
Our experiments using prosodic ABX show that existing S3M representations exhibit sensitivity to prosodic contrasts comparable to that of human listeners.
Moreover, we demonstrate that prosodic ABX is often robust across several experimental conditions, including synthesized speech, in- and out-of-context settings, and different tasks, allowing suitable proxies to be used when data is unavailable.
These findings show that prosodic ABX provides a simple and practical framework for understanding lexical prosody in S3M representations, serving as a reliable guide for model and layer selection for downstream tasks, such as clustering for prosody-sensitive tokens or pronunciation feedback applications.

% \clearpage

\ifcameraready

\section{Acknowledgments}

{We thank Longfei Yang for his feedback during the preparation of this manuscript. We also thank Rina Katsumata and Kentaro Onda for their suggestions regarding collection of the Japanese dataset.
% \red{Finally, we thank reviewers and ....}}
\else

\fi

{
\section{Generative AI Use Disclosure}
Generative AI was used for minor polishing and review of this manuscript. It was used for code generation, most heavily for dataset normalization, plotting, and table formatting. However, all experiments were conceived by the authors and all final code was manually reviewed and edited for clarity.
}

\bibliographystyle{IEEEtran}
\bibliography{mybib}

\end{document}